\documentclass{article}

\usepackage[preprint]{neurips_2025}

\usepackage[utf8]{inputenc} % allow utf-8 input
\usepackage[T1]{fontenc}    % use 8-bit T1 fonts
\usepackage{hyperref}       % hyperlinks
\usepackage{url}            % simple URL typesetting
\usepackage{booktabs}       % professional-quality tables
\usepackage{amsfonts}       % blackboard math symbols
\usepackage{nicefrac}       % compact symbols for 1/2, etc.
\usepackage{microtype}      % microtypography
\usepackage{xcolor}         % colors

\title{Advocate for Complete Benchmarks for Formal Reasoning with Formal/Informal Statements\\ and Formal/Informal Proofs}

\author{%
  Roozbeh Yousefzadeh \\%\thanks{Use footnote for providing further information
  Huawei Hong Kong Research Center\\
  \texttt{roozbeh.yz@gmail.com} \\
  \And
  Xuenan Cao \\
  Department of Cultural Studies \\
  The Chinese University of Hong Kong \\
  \texttt{xuenancao@cuhk.edu.hk} \\
}

\begin{document}

\maketitle

\begin{abstract}
    This position paper provides a critical but constructive discussion of current practices in benchmarking and evaluative practices in the field of formal reasoning and automated theorem proving. We take the position that open code, open data, and benchmarks that are complete and error-free will accelerate progress in this field. We identify practices that create barriers to contributing to this field and suggest ways to remove them. We also discuss some of the practices that might produce misleading evaluative information. We aim to create discussions that bring together people from various groups contributing to automated theorem proving, autoformalization, and informal reasoning.
\end{abstract}

\section{Introduction}  

The field of formal reasoning has emerged as one of the frontiers in the machine learning community~\citep{yang2024formal}. One of the shared goals of this community is to develop AI models that can compete with humans in prestigious mathematical competitions such as the International Math Olympiad (IMO) building on previous achievements of automated systems such as DeepBlue~\citep{pandolfini1997kasparov} and AlphaGo~\citep{silver2017mastering}. This field is also closely related to the field of formal mathematics: a thriving community that writes mathematical proofs using the language of formal verification systems such as Lean~\citep{de2015lean}. Proofs written in Lean are machine-verified. This approach of formalizing mathematics is currently used to digitize existing mathematical theorems as well as to develop new mathematical discoveries~\citep{buzzard2024mathematical}. The machine learning community is in close collaboration with this field, using their formalized theorems to develop AI models, and at the same time, developing models that can help the mathematicians in writing formal proofs \citep{song2023towards}, e.g., \cite{tao2025youtube} using OpenAI's o4-mini model to write a proof in Lean.

Formal reasoning with computers has a long, and continuous literature, before the rise of computers \citep{turing1948intelligent} all the way through current times in fields such as formal verification, computational logic, programming languages, etc. Formal reasoning in its most recent form, in the AI community, emerged after the rise of Large Language Models (LLMs), and specifically, after \cite{polu2020generativelanguagemodelingautomated} showed the remarkable ability of LLMs to learn the language of formal verification systems and write formal proofs, building on earlier work such as by \cite{yang2019learning}. The miniF2F benchmark \citep{zheng2021minif2f} has allowed a community to come together and make shared progress in this field. While miniF2F was initially focused only on the formal language and Automated Theorem Proving (ATP), it soon was provided in informal language, too, to be used in autoformalization and informal reasoning \citep{wu2022autoformalization,jiang2022draft}. Autoformalization is the translation of informal language to formal language, but in the literature, it usually refers only to translation of problem statements and not the proofs. Writing the proofs using an automated system, whether using the informal proofs or not, is usually known as ATP. There is also the task of in-formalization which is the translation from formal language to an informal one.

In this paper, we discuss some of the challenges in the field of formal reasoning. We explain why certain practices slow down the progress in the field and obscure the evaluation of the progress. To address the issues, we suggest altering the practices regarding benchmarks advocating for an open code, open data approach. This approach is aimed at making it easier for other researchers to contribute to formal reasoning and accelerate the progress.

\subsection{Challenges}

One of the difficulties that the formal reasoning community has been dealing with is the scarcity of high-quality data paired in formal and informal languages. Mathematical data in informal language is abundant, but formal data is scarce and rarely paired with high quality informal language. Mathlib, the mathematical library of Lean, is large but succinct, and there are rarely elaborate natural language descriptions for individual theorems.

Moreover, although a formal verification system, such as Lean, makes verification of proofs automated and straightforward, learning the formal language is not trivial for humans and LLMs. This often makes even the evaluation of the accuracy of autoformalization models difficult, because evaluating the correctness of the formal outputs of the LLMs is not trivial for LLMs and also for a human that is not educated on Lean. As we will report later in the paper, some of the reported accuracies in the literature, automatically evaluated by LLMs, are incorrect. The reported accuracy of the SoTA (State of The Art) autoformalization model \citep{gao2025herald} in the literature is 97\% while its true accuracy is 67\% when evaluated by a Lean expert.

Such difficulty in evaluating the accuracy of models and in measuring progress is present in ATP literature, though to a lesser degree. In ATP, LLMs are in charge of generating the proofs, yet, with a few exceptions, the training sets of these LLMs are not fully revealed. Usually, there is a model that is trained on the Internet data, and then it gets fine-tuned further on some additional data in formal language. That additional data are usually not completely open source. The codes for fine-tuning are often not released, and sometimes even the codes needed to reproduce the evaluation results, e.g., code for search methods. The lack of transparency has led to some other members trying to keep the formal proofs away from the public domain.

The other challenge in this field is the errors and discrepancies in the benchmarks. miniF2F was released in 2021, but even as late as April 2025, new errors and unprovable theorems are being discovered, the latest of them being 5 errors reported by \cite{wang2025kimina}. Taking four years for these errors to be gradually discovered implies, perhaps, a wide-spread lack of attention to verifying the correctness of these theorems. The inattentiveness to verification of the correctness of theorems in the benchmark also reflects an under-appreciation of those who can write formal proofs from scratch. The writing of proofs takes time (a person familiar with Lean can write the formal proofs for all the problems in the dataset in a time frame of several months), but the task of writing proofs is underappreciated when the community is accustomed to taking data from the Internet. From the autoformalization perspective, there still remains widespread discrepancies between formal and informal statements in miniF2F. New benchmarks released in the community also do not include formal proofs. An example is the Putnam Bench~\citep{tsoukalasputnambench}, which calls all users of the benchmark to keep generated formal proofs away from the public domain to avoid contamination.

% While one person familiar with Lean could have written the formal proofs for all these problems in a time frame of a few months. Writing the formal proofs takes time, time that is not often appreciated by some members of the community, members that are often used to taking their data for free from the Internet. 

% New benchmarks being released in the community also do not provide the formal proofs, and even request all users of the benchmark to keep any generated formal proofs away from the public domain, e.g., the Putnam Bench \citep{tsoukalasputnambench}.

\subsection{Summary of our position}

\textbf{We take the position that special attention and effort should be paid to ensure that the benchmarks are complete and error-free, verified by the creators of the benchmarks. A complete benchmark will contain 4 components: formal problem statements, informal problem statements, formal proofs, and informal proofs.} Further discussion is provided in Section~\ref{sec:benchmark}.

% It has taken more than 4 years to discover the errors of miniF2F, and the community is still dealing with those shortcomings. Using this hindsight, we can do better in the future. 

\textbf{We take the position that as much as possible, the community should encourage the release of code and training data for developing autoformalization and theorem provers.} In the computer vision community, for example, releasing codes, models, and data is the general practice while holding back on any of those is rare. We suggest the same approach for the ATP community. If the ATP community increases the transparency about training sets, other members of the community will not need to hide the formal proofs in attempts to prevent training set contamination. The hiding of the proofs, which slows down the field, is caused by the simple fact that most models do not release their training sets. Further discussion is provided in Sections~\ref{sec:barrier} and~\ref{sec:evaluation}.

% In our view, both of these issues can be resolved, if the community starts to measure progress when a paper releases its training set and its training code. 

% \textbf{We further take that training sets of models should be discussed clearly, and, as much as possible, data should be open source. As long as models are not upfront about their training sets, other members of the community will try to hide the formal proofs in an attempt to prevent training set contamination. This practice of hiding the formal proofs slows down the progress in the field} but it is caused by the fact that most models do not reveal their training sets. In our view, both of these issues can be resolved, if the community starts to measure progress only if a paper provides adequate information about its training set and its training code.

% If a computer vision model is trained on some hidden training data and it achieves sota accuracy on this benchmark, this result would not have been praised in the cv community. If that was the norm, then others would have started to come up with new datasets and then hiding part

Finally, \textbf{we take the position that the evaluation of the accuracy of autoformalization models cannot yet be completely outsourced to LLMs when evidence suggests that their estimated accuracies are widely inflated.} Further discussion is provided in Section~\ref{sec:evaluation_autoformalize}. 

% Evaluation of progress in this field should follow the standards of scientific research.

\section{Community practices that slow down progress and create barriers for contributing to the field of formal reasoning} \label{sec:barrier}

The following is an illustration of the challenges that a researcher entering the field of formal reasoning might encounter. Imagine that you are a researcher considering to contribute to the field of formal reasoning. You may have read the news about Alpha Proof~\citep{alphaproof}, and are interested in doing research in this field. There are several LLMs freely available on Hugging Face for the task of theorem proving. Maybe you choose Kimina Prover \citep{wang2025kimina}. Then you aim to reproduce their accuracy results. There is no code released for evaluating their accuracy. If you dig a bit into the literature, you will find the evaluation code provided by Deep Seek Prover v1.5. But that code uses Lean 4.09 while Kimina Prover uses Lean 4.15. After some struggle, you manage to change the Lean version and you succeed in evaluating the accuracy of Kimina Prover on miniF2F. Their work is reproducible.

Maybe you investigate some of the failure cases of Kimina Prover, i.e., the theorems from miniF2F that it has not been able to prove. You look at the theorem statements in Lean and in natural language, read their informal proofs, but it would not be clear to you what would a complete correct proof look like, and how far the generated output of the model is from a correct formal proof for these theorems. This is not trivial. If you manage to get access to someone who knows Lean, they might give you some insights, but without that, you would not even know how long the proof should be for such a problem. Let us consider two problems from the failure cases of Kimina Prover: IMO 1985 P6 and $algebra\_cubrtrp1oncubrtreq3\_rcubp1onrcubeq5778$ ($algebra\_5778$ in short). The manually written formal proof for IMO 1985 P6 has more than 1,200 lines while the proof for $algebra\_5778$ has about 100 lines. When prompting Kimina Prover, it generates a proof with 3 lines for IMO 1985 P6 while its proof for $algebra\_5778$ has 44 lines. Comparing the 3 lines to 1,200 lines, and then comparing 44 to 100 lines, there is no correlation or insights there. Hence, unless you have the complete proofs for these problems, you cannot know for sure how far the output of Kimina is from a correct proof, and therefore, it may not be clear to you how to improve this model.

Maybe you consider looking at a more recent benchmark: Putnam Bench. You look at the GitHub repository and the leader-board webpage of the benchmark, and you see the list of models that have achieved the highest accuracies. But, it is not clear which problems they have been able to prove. To find that, you have to dedicate a few GPUs for a couple of weeks to run your LLM on the entire 657 theorems in this benchmark to finally identify the 10 problems that Kimina Prover can prove given 192 attempts at each problem. This is the only way you can access the list of proved theorems and their proofs. Otherwise, the benchmark requests that all formal proofs to be kept away from the public domain, i.e., another community created barrier.

If you decide to reproduce the results of BFS-Prover \citep{xin2025bfsproverscalablebestfirsttree}, you have to write a code for their search method yourself because they have only released their LLM and not the code for their search method. Unless you write that code yourself, you will not be able to reproduce their accuracy results, and even if you attempt to write a code, you might not implement the search method exactly like them. Because of this, you might consider this work irreproducible.

% Let us assume that you get lucky, and through some mentor, you get your hands on some formal proofs that are closely related to the failure cases of Kimina Prover on miniF2F. You consider fine-tuning the model on such data. 

If you get access to some formal proofs that are closely related to the failure cases of Kimina Prover through other channels, such as privileged connections to a mathematician, you can consider fine-tuning the model on that data. You read the literature and read the data augmentation methods for generating formal data for fine-tuning, but the code is not publicly available. After some effort, you manage to write a code and generate the fine-tuning data, not sure if your generated data is in its best format. Then you want to fine-tune the model, but there is no fine-tuning code released with these theorem provers. The fine-tuning code for Lean-Dojo~\citep{yang2023leandojo} is available, but their model is an encoder-decoder while Kimina Prover and most other provers are decoder-only models. So, you locate some fine-tuning scripts for general-purpose decoder models (perhaps from the Llama factory) and try to modify them to use for your theorem prover. There are many settings and hyper-parameters involved. Hence, you will have to repeat the fine-tuning in many different ways, each time changing the hyper-parameters of your fine-tuning, changing your data augmentation techniques and format, and many other settings, to see if you can improve the accuracy of the model. All of these efforts and experiments might take a few months. All of this can happen, if you have access to high quality data to begin this endeavor. In an alternative approach, you might decide to learn Lean and manually write the proofs for those failure cases. By the time you write those proofs, there is a chance that some other member of the community has done the same and will release those proofs just at the time that you finish.

Now imagine not just one, but several researchers going through a similar path. Consider the amount of parallel redundant work that each researcher has to do just to reproduce the other people's work in the literature. Also, consider the redundancy of various people trying to manually identify errors in miniF2F and correcting those errors, or to manually write the proofs. Many of the barriers, redundancies, and difficulties discussed above are community-made. In the following, we discuss these barriers in more detail and examine possible ways to reduce them.

% Such path does not sound very appealing especially if you do not have 

% For the next step, you might want to consider fine-tuning this model on some additional data. 

\textbf{1. Releasing benchmarks without the formal proofs.} 
The community currently follows the convention of releasing a benchmark with formal statements without the formal/informal proofs and letting others write the formal/informal proofs. Putnam Bench, the recent benchmark that is being widely adopted, also follows this practice. Other recent benchmarks such as~\cite{hu24minictx} and~\cite{yu2025formalmath} also do not make an effort to provide the formal or informal proofs, and it remains unclear whether they contain the kind of errors present in the original version of miniF2F.

In our view, this practice slows down the progress in the field. First, as we have seen in the case of miniF2F, many of the theorems might be wrongly formalized, unprovable, and it might take years for these errors to be discovered, if there is not a dedicated effort to identify such issues.

% many of the theorems might be unprovable, wrongly formalized

% bfs prover attempted a million time on the unprovable theorems

Second, for researchers who are proficient in developing AI models, it should not be a prerequisite to learn a formal language to contribute to this field. Such researchers should be able to have a clear perspective about what the AI model should produce, how the proofs look like, how long they are, how many reasoning steps are required, and how far the current output of a model is from the correct proof. If the correct proofs are at most 10 lines, an AI researcher would choose a different LLM with different settings, such as context length, compared to the case where the LLM is supposed to write a 1,000-line proof. Merely giving a problem statement and a formal verification system that gives a binary output for each proof as to whether it is correct or not will slow the progress and obscure the evaluation. Allowing access to formal proofs will facilitate progress in the field as it lowers barriers to contribute and invites innovative methods to make progress.

% On the other hand, if formal proofs are available, everyone has more insight about how to make progress.

Moreover, until proofs are available, many researchers will attempt to write the proofs manually or in collaboration with the LLMs, and these efforts will be parallel, and largely redundant. Formal reasoning requires some knowledge of the language of a formal verification system such as Lean, Isabelle~\citep{paulson1994isabelle}, Coq~\citep{huet1997coq}. Familiarity with Lean requires dedication and time. And familiarity does not guarantee proficiency is solving hard problems. Being able to write a formal proof for some of the IMO problems may require mathematical education, knowledge of various parts of the mathematical library of the formal language (such as Mathlib for Lean), as well as time and effort. The proof for IMO 1985 problem 6 in the miniF2F consists of about 1,200 lines of Lean code. Writing this proof takes time even if the author is already well familiar with Lean and Mathlib.

Because this effort is considerable and costly, people would seek recognition for their effort. One way to get such recognition is using the data to develop a model that beats the SoTA accuracy. Hence, many researchers will be doing the same work in parallel, while only the largest teams will get the recognition. A small group of researchers will not be able to generate high-quality data and at the same time develop LLMs that can prove theorems. Hence, contributing to the field will have a high barrier.

As a guide, we can consider the case of how computer vision researchers deal with ImageNet~\citep{deng2009imagenet}. ImageNet contains classes such as "Haliaeetus leucocephalus", "Carassius auratus", "Latrodectus mactans". Many members of the computer vision field are unfamiliar with many of the classes in the ImageNet to the extent that they might not have heard of these names ever in their lives. Yet, they have been able to contribute to this field without learning how a "Latrodectus mactans" looks like. Many of the papers in computer vision make incremental yet useful contributions without necessarily beating the SoTA accuracy in each paper.

\textbf{2. Releasing benchmarks with formal theorem statements without properly evaluating the ability of AI models in proving them.} 
The benchmarks for ATP often report the accuracy of a few LLMs on the theorems. However, there have been instances where a benchmark is proposed without evaluating it using the standards of the ATP community, i.e., giving several LLMs at least 32 attempts to write a formal proof. An example is the work by \cite{murphyautoformalizing} which proposes a benchmark for autoformalization, and only evaluates the ability of a general purpose model, GPT-4, to prove the theorems while giving it only 5 attempts.

% by merely providing the formal and informal statements while only evaluating the ability of a LLM in autoformalizing the informal statements without an attempt to prove any of the theorems, for example \cite{}. 

% We argue that all the aforementioned four components are essential parts of a formal reasoning benchmark: informal/formal statements and informal/formal proofs. Any benchmark that does not include all these components can be considered an incomplete benchmark for formal reasoning.% Writing theorems in formal language and saying that they should only be used for autoformalization is not a complete contribution. If some theorems are not meant to be proved, why should they be autoformalized?

\textbf{3. Releasing benchmarks without informal statements and informal proofs.} 
Similarly, any benchmark released in formal language can also be considered a benchmark for autoformalization. If some mathematical theorems are worthy of proving with an automated system, it would be needed at some point to have an automated system that can autoformalize and informalize them. Similarly any theorem worth proving in formal language may have an equivalent proof in natural language since providing the informal proofs can help the formal and informal reasoning \citep{jiang2022draft}.

Hence, releasing any incomplete benchmark creates burden on the community as others will have to complete the missing pieces without getting credit. This can lead to the missing pieces being completed with a lower quality since no credit is earned, and also, since people who complete the benchmark may not be well familiar with it. This has been the case for miniF2F which still contains considerable discrepancies between its formal and informal statements.

% \subsection{Not fixing the shortcomings of datasets for the public}

\textbf{4. Not releasing code for fine-tuning and/or evaluation.} 
Often, merely one LLM is released for each new contribution to the field of ATP. Even in the cases where the evaluation relies on a specific search method separate from the LLM, some times that code is not released, making it rather difficult to reproduce the results or to build on the method. For example, BFS-Prover only releases its LLM, but does not release the code for its search method.

For the LLMs that do whole proof generation, fine-tuning script is often not released. Examples include Kimina Prover, BFS Prover, Goedel Prover~\citep{lin2025goedelproverfrontiermodelopensource}, DeepSeek Prover. There is also the case of Alpha Proof which reached the level of silver-medal at IMO 2024, but only released the proofs for the problems along with a blog post without even releasing a technical paper describing the method.

% There are models in the industry that are closed source and they also do not seek recognition from the scientific research community seeking to publish a paper, e.g., OpenAI's models. However, seek to be published as a scientific contribution yet they do not release their code.

\textbf{5. Concealing the proofs from others.} 
miniF2F was originally open about adding the proofs to its Github repository as the proofs were becoming available. However, at some point, it stopped accepting pull requests that provided new proofs. Moreover, Putnam Bench explicitly asks its users to keep the proofs private. This is another community-created barrier that is, fortunately, sometimes not observed by all. For example, Kimina Prover recently released formal proofs for some of the unproved problems in miniF2F without specifying whether the proofs were written for the first time manually or by the LLM. Abiding by the request of Putnam Bench creators, Kimina Prover refrained from releasing the proofs for the 10 theorems that it had proved from that benchmark.

\section{Our position about benchmarks} \label{sec:benchmark}

\textbf{1. We advocate new ATP benchmarks to be complete.} 
As described earlier, we consider a formal reasoning benchmark complete if it contains four components: informal statements, formal statements, informal proofs, formal proofs. We suggest that the community advocates for any new proposed benchmark to be complete. This proposal will make benchmark contributions a more challenging task, but it will lead to higher quality benchmarks.
%First, requiring the creators of a benchmark to be responsible to create all the four components may lead to higher quality and a better match between the four components.

The creators of a benchmark will be welcome to use any automated or manual tools to create the benchmark and write the formal proofs, but at the end, they will be responsible for the correctness of all four pieces of the benchmark. Specifically, requiring the formal proofs to be written before the release of a benchmark ensures that all theorems are provable and correct.

The inclusion of informal statements and informal proofs facilitates research leveraging informal reasoning and informal proofs for formal reasoning. When these two pieces are put together after the release of a benchmark, often, it has led to lower quality match between the formal statements and the informal ones as is the case for miniF2F.

Moreover, when a benchmark is released without some of the four pieces, it leads to parallel efforts. Parallel work and effort are inherent parts of research and in most cases unavoidable because it is hard to predict in advance. However, incomplete benchmark poses a certain yet avoidable burden of a different kind: it implicitly shifts the completion task from the benchmark creators to communities which may or may not have adequate resources and incentives to achieve high qualities.

% However, when an incomplete benchmark is released, the completion task is implicitly put on the shoulders of the community, and some people will eventually take this task upon themselves, most likely, simultaneously with others.

We also note that some research groups in the community do not have enough familiarity with the language of formal verification systems needed to perform human verification, correct the errors in the datasets, or complete a dataset by adding the missing pieces such as informal proofs/statements. As a result, the barrier for such groups to use an incomplete benchmark will be higher. For example, \cite{xin2024deepseek} showcases a formal proof written by its model for a shortlisted IMO problem without realizing that the formal problem statement is wrongly formalized (including a false statement and the proof leverages that false statement to prove False). Not realizing this mistake and even showcasing it as an achievement is indicative of the lack of familiarity with Lean and also the basic mathematics involved in proving the theorem. Most recently, BFS-prover \citep{xin2025bfsproverscalablebestfirsttree} released an IMO proof that ends with the tactic \textit{swap} which is applicable when a proof has at least two unproved goals. A proof ending with swap is an incomplete proof.

We note that in the field of formal reasoning, benchmarks are not only about autoformalization and ATP. A more recent line of research is focused on conjectures, i.e., developing AI models that compose new theorems that might be useful, e.g., \citep{formalconjectures2025}. This line of research primarily deals with conjectures and exploration of possible mathematical definitions and theorems. By definition, this line of research has to deal with many conjectures that may be unprovable. For the provable conjectures, it remains to be seen whether they are useful or not. This line of research can possibly generate ATP benchmarks down the line, but its immediate focus is not that. Completeness, as we defined for ATP benchmarks would not be meaningful in such a setting, and we do not advocate for such benchmarks to be complete.

% As a hypothetical, consider a benchmark that does not include informal statements and informal proofs. Now, consider a research group with experience in informal reasoning that is interested to use this benchmark but does not have any expert in the formal language. Such group would not be able to use this benchmark easily. They might make the effort to find some equivalent informal statements on the Internet or in text books, but they would not be able to ensure that those informal statements are exact translation of the formal statements in the benchmark. Such group may actually proceed with assembling a set of informal statements to complete the benchmark, and perhaps achieve some accuracy gains as well, but because of their lack of expertise in the formal language, the quality of the assembled dataset may be low. This is a hypothetical, but it may remind us of the widespread discrepancies between the formal and informal statements in the miniF2F.

\textbf{2. We need to make an effort to complete the existing benchmarks.} 
For any existing benchmark that is not complete, making the effort to complete them will facilitate progress. Existing benchmarks were used in times when challenges were different. When the creators of the miniF2F first released this benchmark, it was meant to be a benchmark only for automated theorem proving in formal language. Initially, the field of autoformalization did not have many active researchers, and papers that leverage informal proofs to aid formal reasoning \citep{jiang2022draft,wang2024legoprover} were not written yet. In such context, it was not clear to the community how helpful it would be to include the informal statements and proofs, especially those that would exactly match the formal ones.

Over the years, many mistakes were discovered in the formal statements of miniF2F. This was a gradual process where one paper would report and fix a few errors, and another paper would report and fix a few other errors, which is still the continuing trend. It is not uncommon for benchmarks to include some errors or shortcomings as reported by \cite{vendrow2024large}. However, in the field of formal reasoning, it is more straightforward to detect and fix those errors when one attempts to write formal proofs for the theorems in a benchmark. Considering that one person familiar with Lean could write the formal proofs for all these theorems in a time frame of a few months and ensure their correctness, taking the community four years to gradually discover and fix these errors indicate the inefficiency of the current approach.

Another factor contributes to this lack of effort/attention to the errors in this dataset. There is a competition in the community, rightly so, to improve the SoTA accuracy on these benchmarks, yet when reporting an improvement in the accuracy, some members would prefer to show that no manual effort was involved in writing the proofs. Usually, the story in the paper construct the following narrative: we acquired new data (that we may partially share), then we trained a LLM on that data (using a code that we are not going to release), and that process led to a better accuracy. Involving a human in the process even to verify the correctness of formal statements and provability of the theorems would seem to lessen the achievement (because it would imply the work is "semi-manual" instead of "fully automated"). It took some time for some members of the community to get past this view and realize that some of the theorems in miniF2F are actually unprovable and changing them is the only way forward.
     
We believe not having access to the formal proofs is an unnecessary distraction. Having access to the proofs is necessary and helpful to advance the field of formal reasoning. It would be helpful to note that many of the contributors to the computer vision field have no idea what is a "Latrodectus mactans", one of the classes in ImageNet, yet those same people have been able to make great contributions to developing AI models that can classify images of "Latrodectus mactans" from other images or to detect a "Latrodectus mactans" inside an image and draw a boundary around it. The computer vision community has made progress in all aspects of AI development and deployment partially due to open access to the entire test sets of benchmarks such as ImageNet along with the labels. Having access to the labels of ImageNet has been helpful for the ML community without hindering the progress in the generalization abilities of the models \citep{recht2019imagenet}.

\textbf{3. We need to make an effort to cross check the match between formal and informal statements in existing benchmarks, and correct any mismatches.} 
There are benchmarks that appear to be semi-complete such as miniF2F, but the quality of match between the components is low, e.g., there are discrepancies between informal and formal statements for about half of the theorems in miniF2F. For such cases, the community needs to make the effort to correct them. This will help both in better performance of the models, and it will also lead to better evaluation of progress in the field. For example, consider the theorem $aime\_1987\_p5$ from miniF2F which Kimina Prover 7B cannot prove pass 32. When we corrected the informal proof for this problem, the model was able to prove the theorem pass 1. So, higher quality benchmarks can be helpful for everybody.

% \subsection{When a formal proof becomes available, do not advocate for it to be kept away from the public domain}

% \subsection{Suggestion 2: Include the formal proofs in the benchmarks as much as possible}

\textbf{4. We must credit manual efforts for all the items above.} 
To facilitate the progress in the past three items, the community can use all its resources, including manual efforts of its members. For this engagement to thrive, it would help if the machine learning community give value and credit to such efforts even when they are manual or semi-manual.

It would be helpful to note that languages such as Lean are also programming languages, the same way that Python is a programming language. Coding in Lean is not necessarily easier or harder than coding in Python. Therefore, when someone writes a proof in Lean, it can be considered manual labor the same way that writing a PyTorch script for fine-tuning a LLM is manual. From this perspective, it would be easier for some members of ML community to give credit to the efforts behind writing formal proofs. To develop AI models that can do formal reasoning, we need formal proofs as much as we need PyTorch scripts for fine-tuning and evaluation of LLMs.

% \subsection{Suggestion: When a formal proof becomes available, do not advocate for it to be kept away from others}

% \subsection{Verify formal statements are correct translation of original problems}

% \subsection{Verify the correctness of all }

% give credit to and value the manual effort for correcting and completing the benchmarks

% for example, see the comments of reviewer \#2 for the paper by \cite{} that provides the proofs for IMO problems in miniF2F. Reviewer \#2 considers the work manual, and therefore, lacking substance \citep{}.

\section{Our position about evaluation methods}  \label{sec:evaluation}

\subsection{Automated Theorem Proving} \label{sec:evaluation_ATP}

\textbf{1. Release training data and perform ablation studies} 
It has become a common practice that contributions to the field of formal reasoning do not release training data or  report ablation studies to reveal the reason behind their gained accuracies. Sometimes, model creators reveal that they have benefited from human annotators and evaluators, e.g., \citep{wang2025kimina}. But often it remains unclear how human labor have been involved in the process and to what extent they have helped. Have they written any complete formal proofs for any of the problems? When a paper comes forward revealing that humans were involved in the process, we believe this should be welcomed by the community. However, sufficient information should be given so that experiments are reproducible.

Many other models do not clearly reveal human involvement, yet reading between the lines it becomes clear that they have benefited from such assistance. For example, Deep Seek Prover v1.5 \citep{xin2024deepseekproverv15harnessingproofassistant} in one sentence mentions that its model was trained on the validation set of miniF2F. Yet, it never releases those proofs. And it also does not mention whether all the proofs of validation set was formalized and used in its fine-tuning or only part of the validation set was used. In addition, it does not report its accuracy on the validation set. When we evaluated the accuracy of this model on the miniF2F validation set, we obtained an accuracy of 64\% pass 32. Given the information provided by the authors, it remains unclear why the model does not achieve near-perfect accuracy on the miniF2F validation set. Did the authors only train on 64\% of the miniF2F validation set? What was the training loss at the end of fine-tuning? For the 36\% of the theorems that it cannot prove, has the model seen the proofs but cannot reproduce them? Without a clear statement by the authors, the interpretation of the accuracy of the model remains within a fog of questions.

Yet, with all these questions hanging about the model of Deep Seek Prover v1.5, another group of researchers used that very same model and fine-tuned it on some additional data to obtain the Goedel Prover, a new model which was the new SoTA at the time of its release. The creators of Goedel Prover also did not release their additional training set, nor the code they used for additional fine-tuning. These practices make the contribution of the work unclear. It is not clear what exactly is the driving force behind the accuracy obtained. Is it the data? Is it the fine-tuning method? In general, there are usually no ablation studies of the improvements achieved by each new model.

% We suggest that models reveal their training data and also perform ablation studies to provide insights about the source of their achievements.

\textbf{2. Report insights beyond accuracy, such as proof lengths.} 
Often, only one accuracy metric is reported without providing additional insight about the proofs. Some of the theorems in miniF2F can be proved outright using automatic solvers in Lean, e.g., nlinarith, linarith, omega. One theorem in Putnam Bench can be proved using only the linarith tactic. It is never reported how many theorems of a benchmark are proved using those solvers alone, and all the credit is given to the LLMs.

Moreover, insights about the length of the proofs are not provided in the literature. Goedel Prover which used to have the SoTA accuracy on miniF2F in early 2025, could only prove theorems with proofs as long as about 10 lines. This information was never shared in the literature, neither by Goedel Prover nor by other models. Kimina Prover, however, is able to write elaborate formal proofs as long as a few hundred lines. This sets Kimina Prover above previous provers, but it appears that this achievement is made possible not just by better training data but also by increasing the context length of the LLM.

It would be crucial for the community to start reporting more information about their achievements beyond a simple accuracy metric and to make fair comparison with other models as well. Reporting a simple accuracy metric does not tell the full story about the capabilities of a model and the path to its achievements. We suggest, at the minimum, that the lengths of the proofs be reported, as well as the number of theorems that are proved only by automatic solvers. This suggestion is not about replacing the accuracy metric with other metrics, rather, it's about keeping the accuracy metric while providing more elaborate interpretation of the correct automatically written proofs.

% not reporting the full story while cherry picking some

% report the role of automatic solvers inside Lean

% just a mere accuracy metric - not mentioning the number of lines in the proofs

\textbf{3. Consider full pipelines of autoformalization models plus theorem provers.} 
The staring point for ATP models is usually formal statements in a benchmark. However, this implies presence of a human in the loop that has translated the problem from informal language to formal because the formal statements in the benchmarks are written by humans. A more realistic pipeline would consider an AI system that starts from the informal language, translates the theorem to the formal language, and proves the theorem in the formal language. And finally, the proof is compared to the original theorem statement. Recently, \cite{xuejun2025mathesis} has considered a pipeline where the starting point is the informal statement which is a step towards the above pipeline, yet, their pipeline is not complete as it does not explicitly compare the formalized statements against the original NL statements. As a result, if a model simplifies the theorem during the formalization, and the ATP model proves such a simplified theorem, the pipeline receives the full credit for proving the problem. This approach implicitly rewards simplification of problems by autoformalizers because simplification makes both tasks of autoformalization and ATP easier. However, in the real world, this approach would not get full credit, the same way that when a human participant of IMO writes a proof for a simplified version of a problem, they will not get a high score for their effort. We will discuss the motivation for such automated pipeline later in Appendix~\ref{sec:discovery}.

% some members of the community appear worried to report that 

% Finding errors in the 1,000,000 images of ImageNet is not an easy task, especially considering the 1,000 labels. However, a person familiar with Lean can write the formal proofs for the problems in miniF2F

\subsection{Autoformalization} \label{sec:evaluation_autoformalize}

% There are two major difficulties in this field. 

% \textbf{Suggestion 2: Human verify the match between formal/informal statements of the benchmarks that are widely used}

\textbf{1. Release training data and perform ablation studies.} 
This position is similar to the position described for ATPs. When releasing a new LLM that achieves a higher accuracy, it would be necessary to release the training sources and training scripts. If the training method is different from the previous methods, an ablation study would be necessary to reveal the effect of training data vs. the effect of training method.

\textbf{2. Avoid using unverified automated pipelines to evaluate the accuracy of autoformalization models. Verify the accuracy of such automated pipelines before deploying them in scientific research or practice.} 
In autoformalization, there is no automated system that can verify the correctness of a translation with high certainty. This is quite different from the case of evaluation for ATPs where the formal verification system can verify the correctness of a proof with absolute certainty. Still, the formal verification system can help automatically detect any semantic errors in a formalized statement, but when there are no such errors, it does not mean that the translation is correct. In fact, in many instances, LLMs can write seemingly plausible formal statements that could take time for a human expert to detect their errors.

Another difficulty is that a single informal statement can have several different correct translations in a formal language. If one has a correct translation of an informal statement (i.e., ground truth in benchmark), merely comparing the output with ground truth may not lead to correct evaluation of the accuracy of the LLM's output. Some papers have focused on automating the evaluation of the equivalence of formal statements with some success \citep{li2024autoformalize, murphyautoformalizing}, but they are still far from being reliably deployed in practice.

Another more recent evaluation method is to translate the formal statements back to informal language using a second LLM, and then use a third LLM judge to compare the result with the original statement. If the LLM judge decides that the two informal statements are equivalent, the translation of the formal statement is considered correct. This is the automated evaluation method used by Herald, published at ICLR 2025, reporting the SoTA accuracy of 97\% on miniF2F. We reproduced their pipeline reaching the same accuracy of 97\% with their evaluation method. However, when we manually checked the accuracy of the translations, the accuracy was 67\%, a huge discrepancy with the accuracy evaluated by their LLM judge.

Our position here is that the use of automated systems, such as a LLM judge, to evaluate the accuracy of a model is not problematic in principle. However, before using such system and putting those results in the literature, and claiming SoTA accuracy, one has to first establish the accuracy of such automated systems. An automated system that reports an accuracy of 97\% while the true accuracy is only 67\% is not suitable to be deployed in practice.

\textbf{3. Consider more realistic settings in evaluation where the ground truth is not given to the model.} 
Although the automated pipeline used by Herald, which deploys an LLM as a judge, does not have an acceptable accuracy, the approach remains interesting. There may come LLMs that are reliable in evaluating the correctness of a formal translation by comparing two statements in formal language. This can be one of the main goals that can be pursued in this field. It would be a useful contribution to the field to develop models that can reliably verify the correctness of formal translations merely by comparing the formal and informal statements without the need for the ground-truth translation.

% Moreover, Herald's pipeline relies on the ground truth statement, however, in practice, ground truth translations are not always available. It would be a useful contribution to the field to develop models that can verify the correctness of formal translations without the need to access the ground truth.

\section{Alternative views}

One of the alternative views to our position is that keeping the formal proofs away from the public domain leads to better evaluation practices because once a formal proof is public, it is nearly impossible to prevent it from being used in the training process of LLMs. This view largely shares the same concerns that we expressed about training set contamination and evaluation, however, it appears to be hopeless about people holding back on training the models on such data, so as the remedy, suggests keeping the formal proofs of the benchmarks private. Earlier in the paper, we explained why this approach slows progress and creates an unnecessary barrier to contributing to the field of formal reasoning.

Another alternative view is about training set contamination. Some do not concern themselves with creating a separation between training and testing sets. Any data that they can access are considered fair to be used in model training, and they consider the result useful as long as it improves the accuracy of a model on some benchmark. We do not disagree with this view, as long as they are transparent about the process and the achieved result. Through experiments in the community, it has become clear that many LLMs are not capable of reproducing the contents that they have been exposed to during their training. Hence, being able to train a model so that it can reproduce what it has seen before can be a useful exercise in itself. Such a model might be useful to be deployed in practice, and, moreover, the methods for training such models may also be valuable. If the complete proofs of all the theorems in miniF2F and Putnam Bench are released, it would not be trivial to train a model on those proofs in a way that the trained model can correctly reproduce all those proofs. This can be one of the steps in a sound scientific approach to produce advanced provers. Such step would be aiming to evaluate what models and what training methods are required at the minimum to reproduce such proofs, a question currently without an answer in the literature.

There exists another alternative view suggesting that releasing the proofs for a benchmark will devalue it. After the release of the proofs of IMO problems in miniF2F by \cite{yousefzadeh2025a}, concerns were raised that this could lead to a devaluation of miniF2F. A chain of assumptions lead up to this view: when the proofs of theorems in a benchmark become available, it is trivial to fine-tune a LLM on them so that it can achieve high accuracy on those theorems, which means the SoTA accuracy on the benchmark will soon saturate, and people would want to move onto other more challenging benchmarks. We do not agree with these assumptions. Instead, we have two arguments against it. First, as we discussed above, training a LLM on formal data such that the LLM can correctly reproduce the proofs is not a trivial task, and even after the release of the proofs of IMO problems in miniF2F, most of those problems remain unproved by the best provers. Even with high quality data, achieving high accuracy for ATP is difficult. Second, the ML community is likely to get past the point of not releasing training sources, and the line of research that uses open source models and open source data is likely to grow. This line of research currently has SoTA accuracy of about 40\% \citep{lample2022hypertree} on miniF2F and it is likely to keep a clear separation between its training and test sets. LeanDojo \citep{yang2023leandojo} also has a method to properly divide a set of theorems into a train and a test set.

% mature enough to get past the point of not revealing the training sources. Part of the reason that community is still not upfront about training sources is that generation of high quality data is costly, and even with the high quality data, achieving high accuracy even using high quality data is difficult. In our view, once all the proofs become available for a benchmark, new papers will be written that start with LLMs that are trained on clear training sources with no contamination of the problems in test sets.

% This view might be idealistic but it has already happened in the computer vision community. Despite the availability of test sets with their labels, the community was able to make progress in developing better and better models while keeping a separation between train and test sets. In that community, also, concerns were raised that some of the models might not have generalization abilities, but an independent research demonstrated otherwise \citep{}.

\section{Conclusions}

In this paper, we argue for the field of formal reasoning to adopt a more open approach in reporting its achievements (being upfront about training sets, training methods, etc), and also suggested a more open approach in sharing the achievements within the community (sharing code, sharing data). We also take the position that benchmarks should be complete. We describe how incomplete benchmarks slow down progress in the field and create barriers for others to contribute to this field.

\bibliographystyle{unsrtnat}
\bibliography{refs}

\clearpage

\appendix

\section{Autonomous AI systems excelling in reasoning and scientific discovery} \label{sec:discovery}

With the rise of generative models, we have seen their success in scientific discoveries \citep{wang2023scientific}. For example, FunSearch \citep{romera2023mathematical} succeeded in writing a bin-packing algorithm that is faster than any human written algorithm. These systems, usually consist of a LLM at their core, and they have shown a remarkable ability in automated reasoning. For example, AlphaGeometry \citep{trinh2024solving} was able to reach gold-medal level in solving geometry problems at IMO. AlphaProof~\citep{alphaproof}, similarly reached silver-medal level in proving IMO problems in number theory and algebra. Other examples include AlphaCode~\citep{li2022competition} and AlphaEvolve~\citep{cui2021alphaevolve}.

% These AI systems have also shown remarkable abilities in tasks that were previously considered too specialized to be performed reliably by an AI system.

\cite{silver2025welcome} suggest that we will see a new generation of AI agents that will reach unprecedented abilities predominantly by learning from experience. This argument largely draws from not just recent successes of AI systems, but also the successes of systems such as AlphaGo which was able to learn the game of Go merely by playing with an automated adversary, and ultimately reaching the level of expertise to beat the human champion. When the same algorithm was transformed to play Japanese chess, it also passed the best human performance while the model developers were not familiar with Japanese chess. Indeed, some of the strategies that AlphaGo used were known to fail by human champions, e.g., the famous "fifth line" strategy. Yet, AlphaGo devised those strategies in ways that were able to beat the champions. If the champions had tutored AlphaGo, the model might have learned to not use those strategies, yet, in the absence of such tutoring, it considered those strategies, and developed them in such a way that could be advantageous.

The idea here is that to make new discoveries or to arrive at models that can find better ways of playing a game or excel at proving mathematical theorems, the models have to be given the space to explore the possibilities by themselves. In these applications, there are no negative societal consequences. If a model considers a bad strategy at a game, it will lose the game. Or, if a model, considers a bad strategy in proving a mathematical theorem, or in defining a new theorem, in the worst case scenario, it will fail to prove the theorem. In this line of work, AI models are used for exploration of possibilities as well. The FunSearch method that came up with a new bin-packing algorithm produced 10 million possible algorithms while only a small fraction of them could be considered a correct algorithm. Still, from the small fraction of algorithms that could vbe considered correct, majority of them algorithms were slow compared to existing algorithms, and hence, useless. However, one of those algorithms turned out faster than all the previously available algorithms for this task. Writing a bad bin-packing algorithm does not have any negative societal impact and the correctness and speed of any such algorithm can be easily verified automatically by a Python script.

From this perspective, an automated pipeline for mathematical reasoning that can start from informal statements, translate to a formal language, and proceed with proving, would be preferable to a pipeline that needs a human in the loop for formalizing the statements, verifying the correctness of statements, etc. The human would still be needed to initiate and oversee the process.% and at the end of the formal reasoning pipeline to compare the proof against the original informal theorem statement.

%how about this? I removed 'in the loop'

%I would say "reasoning would be preferable to a pipeline that needs a human for formalizing the statements, verifying the correctness of statements"

%I am about to enter home. Now i am right outside the door.

% Hence, in this work, we suggest a fully automated pipeline to evaluate AI systems on the task of formal reasoning.

\end{document}